\documentclass{article}

\PassOptionsToPackage{numbers, compress}{natbib}

\usepackage[final]{neurips_2019}  

\usepackage{microtype}
\usepackage{graphicx}
\usepackage{subcaption}
\usepackage{booktabs} 

\usepackage{bbm}

\usepackage{color}
\usepackage{multirow}

\usepackage[utf8]{inputenc}
\usepackage[T1]{fontenc}

\usepackage{amssymb}
\usepackage{amsmath,bm}
\usepackage{enumitem}
\newtheorem{theorem}{Theorem}

\newtheorem{lemma}{Lemma}

\usepackage{pifont}

\usepackage{hyperref}

\title{How to Initialize your Network? \\Robust Initialization for WeightNorm \& ResNets}

\author{
  Devansh Arpit\thanks{Equal contribution. Work done while Víctor Campos was an intern at Salesforce Research.} \ \raisebox{2.5pt}{\small \dag}, 
  Víctor Campos\footnotemark[1] \ \raisebox{2.5pt}{\small \ddag}, 
  Yoshua Bengio\raisebox{2.5pt}{\small \S}  \\
  \raisebox{2.5pt}{\small \dag}Salesforce Research, \raisebox{2.5pt}{\small \ddag}Barcelona Supercomputing Center, \\ \raisebox{2.5pt}{\small \S}Montr\'eal Institute for Learning Algorithms, Universit\'e de Montr\'eal, CIFAR Senior Fellow\\
  \texttt{devansharpit@gmail.com}, \texttt{victor.campos@bsc.es}
  }

\begin{document}

\maketitle

\begin{abstract}
Residual networks (ResNet) and weight normalization play an important role in various deep learning applications. However, parameter initialization strategies have not been studied previously for weight normalized networks and, in practice, initialization methods designed for un-normalized networks are used as a proxy. Similarly, initialization for ResNets have also been studied for un-normalized networks and often under simplified settings ignoring the shortcut connection. To address these issues, we propose a novel parameter initialization strategy that avoids explosion/vanishment of information across layers for weight normalized networks with and without residual connections. The proposed strategy is based on a theoretical analysis using mean field approximation. We run over 2,500 experiments and evaluate our proposal on image datasets showing that the proposed initialization outperforms existing initialization methods in terms of generalization performance, robustness to hyper-parameter values and variance between seeds, especially when networks get deeper in which case existing methods fail to even start training. Finally, we show that using our initialization in conjunction with learning rate warmup is able to reduce the gap between the performance of weight normalized and batch normalized networks.
\end{abstract}

\section{Introduction}
\label{sec_intro}
Parameter initialization is an important aspect of deep network optimization and plays a crucial role in determining the quality of the final model. In order for deep networks to learn successfully using gradient descent based methods, information must flow smoothly in both forward and backward directions \citep{glorot2010understanding,he2016deep,hanin2018start,hanin2018neural}. Too large or too small parameter scale leads to information exploding or vanishing across hidden layers in both directions. This could lead to loss being stuck at initialization or quickly diverging at the beginning of training. Beyond these characteristics near the point of initialization itself, we argue that the choice of initialization also has an impact on the final generalization performance. This non-trivial relationship between initialization and final performance emerges because \textit{good} initializations allow the use of \textit{larger} learning rates which have been shown in existing literature to correlate with better generalization \citep{jastrzkebski2017three,smith2018bayesian,smith2017don}.

Weight normalization~\cite{salimans2016weight} accelerates convergence of stochastic gradient descent optimization by re-parameterizing weight vectors in neural networks.
However, previous works have not studied initialization strategies for weight normalization and it is a common practice to use initialization schemes designed for un-normalized networks as a proxy. We study initialization conditions for weight normalized ReLU networks, and propose a new initialization strategy for both plain and residual architectures. 


The main contribution of this work is the theoretical derivation of a novel initialization strategy for weight normalized ReLU networks, with and without residual connections, that prevents information flow from exploding/vanishing in forward and backward pass. Extensive experimental evaluation shows that the proposed initialization increases robustness to network depth, choice of hyper-parameters and seed. When combining the proposed initialization with learning rate warmup, we are able to use learning rates as large as the ones used with batch normalization~\cite{ioffe2015batch} and significantly reduce the generalization gap between weight and batch normalized networks reported in the literature~\cite{gitman2017comparison,shang2017exploring}. Further analysis reveals that our proposal initializes networks in regions of the parameter space that have low curvature, thus allowing the use of large learning rates which are known to correlate with better generalization \citep{jastrzkebski2017three,smith2018bayesian,smith2017don}.

\section{Background and Existing Work}
\label{sec_existing_work}
\textbf{Weight Normalization:} previous works have considered re-parameterizations that normalize weights in neural networks as means to accelerate convergence. In \citet{arpit2016normalization}, the pre- and post-activations are scaled/summed with constants depending on the activation function, ensuring that the hidden activations have 0 mean and unit variance, especially at initialization. 
However, their work makes assumptions on the distribution of input and pre-activations of the hidden layers in order to make these guarantees. 
Weight normalization~\cite{salimans2016weight} is a simpler alternative, and the authors propose to use a data-dependent initialization \citep{mishkin2015all,krahenbuhl2015data} for the introduced re-parameterization. 
This operation improves the flow of information, but its dependence on statistics computed from a batch of data may make it sensitive to the samples used to estimate the initial values.


\textbf{Residual Network Architecture:} residual networks (ResNets)~\cite{he2016deep} have become a cornerstone of deep learning due to their state-of-the-art performance in various applications. 
However, when using residual networks with weight normalization instead of batch normalization~\cite{ioffe2015batch}, they have been shown to have significantly worse generalization performance. For instance, \citet{gitman2017comparison} and \citet{shang2017exploring} have shown that ResNets with weight normalization suffer from severe over-fitting and have concluded that batch normalization has an implicit regularization effect. 


\textbf{Initialization strategies:} there exists extensive literature studying initialization schemes for un-normalized plain networks (c.f.~\citet{glorot2010understanding,he2015delving,saxe2013exact,poole2016exponential,pennington2017resurrecting,pennington2018emergence}, to name some of the most prominent ones). 
Similarly, previous works have studied initialization strategies for un-normalized ResNets~\cite{hanin2018start, taki2017deep,tarnowski2018dynamical}, but they lack large scale experiments demonstrating the effectiveness of the proposed approaches and consider a \textit{simplified ResNet} setup where shortcut connections are ignored, even though they play an important role~\cite{jastrzkebski2017residual}. 
\citet{zhang2019fixup} propose an initialization scheme for un-normalized ResNets which involves initializing the different types of layers individually using carefully designed schemes. They provide large scale experiments on various datasets, and show that the generalization gap between batch normalized ResNets and un-normalized ResNets can be reduced when using their initialization along with additional domain-specific regularization techniques like cutout~\cite{devries2017improved} and mixup~\cite{zhang2018mixup}.
All the aforementioned works consider un-normalized networks and, to the best of our knowledge, there has been no formal analysis of initialization strategies for weight normalized networks that allow a smooth flow of information in the forward and backward pass.


\section{Weight Normalized ReLU Networks}
\label{sec_normed_deep_relu}

We derive initialization schemes for weight normalized networks in the asymptotic setting where network width tends to infinity, similarly to previous analysis for un-normalized networks~\cite{glorot2010understanding,he2016deep}. 
We define an $L$ layer weight normalized ReLU network $f_{\mathbf{\theta}}(\mathbf{x}) = \mathbf{h}^L$ recursively, where the $l^{th}$ hidden layer's activation is given by,
\begin{align}
\label{eq_norm_relu_layer}
    \mathbf{h}^l &:= ReLU(\mathbf{a}^l) \nonumber\\
    \mathbf{a}^l &:= \mathbf{g}^l \odot \hat{\mathbf{W}}^l \mathbf{h}^{l-1} + \mathbf{b}^l \mspace{20mu} l \in \{ 1,2, \cdots L \}
\end{align}
where $\mathbf{a}^l$ are the pre-activations, $\mathbf{h}^l \in \mathbb{R}^{n_l}$ are the hidden activations, $\mathbf{h}^o = \mathbf{x}$ is the input to the network, $\mathbf{W}^l \in \mathbb{R}^{n_l \times n_{l-1}}$ are the weight matrices, $\mathbf{b} \in \mathbb{R}^{n_l}$ are the bias vectors, and $\mathbf{g}^l \in \mathbb{R}^{n_l}$ is a scale factor. We denote the set of all learnable parameters as $\theta = \{ (\mathbf{W}^l,\mathbf{g}^l,\mathbf{b}^l)\}_{l=1}^L$. Notation $\hat{\mathbf{W}}^l$ implies that each row vector of $\hat{\mathbf{W}}^l$ has unit norm, i.e.,
\begin{align}
    \hat{\mathbf{W}}^l_i = \frac{{\mathbf{W}}^l_i}{\lVert {\mathbf{W}}^l_i \rVert_2 } \mspace{20mu} \forall i
\end{align}
thus $\mathbf{g}^l_i$ controls the norm of each weight vector, whereas $\hat{\mathbf{W}}^l_i$ controls its direction. Finally, we will make use of the notion $\ell(f_{\mathbf{\theta}}(\mathbf{x}), \mathbf{y})$ to represent a differentiable loss function over the output of the network.

\begin{figure}
\centering
  \includegraphics[width=\linewidth]{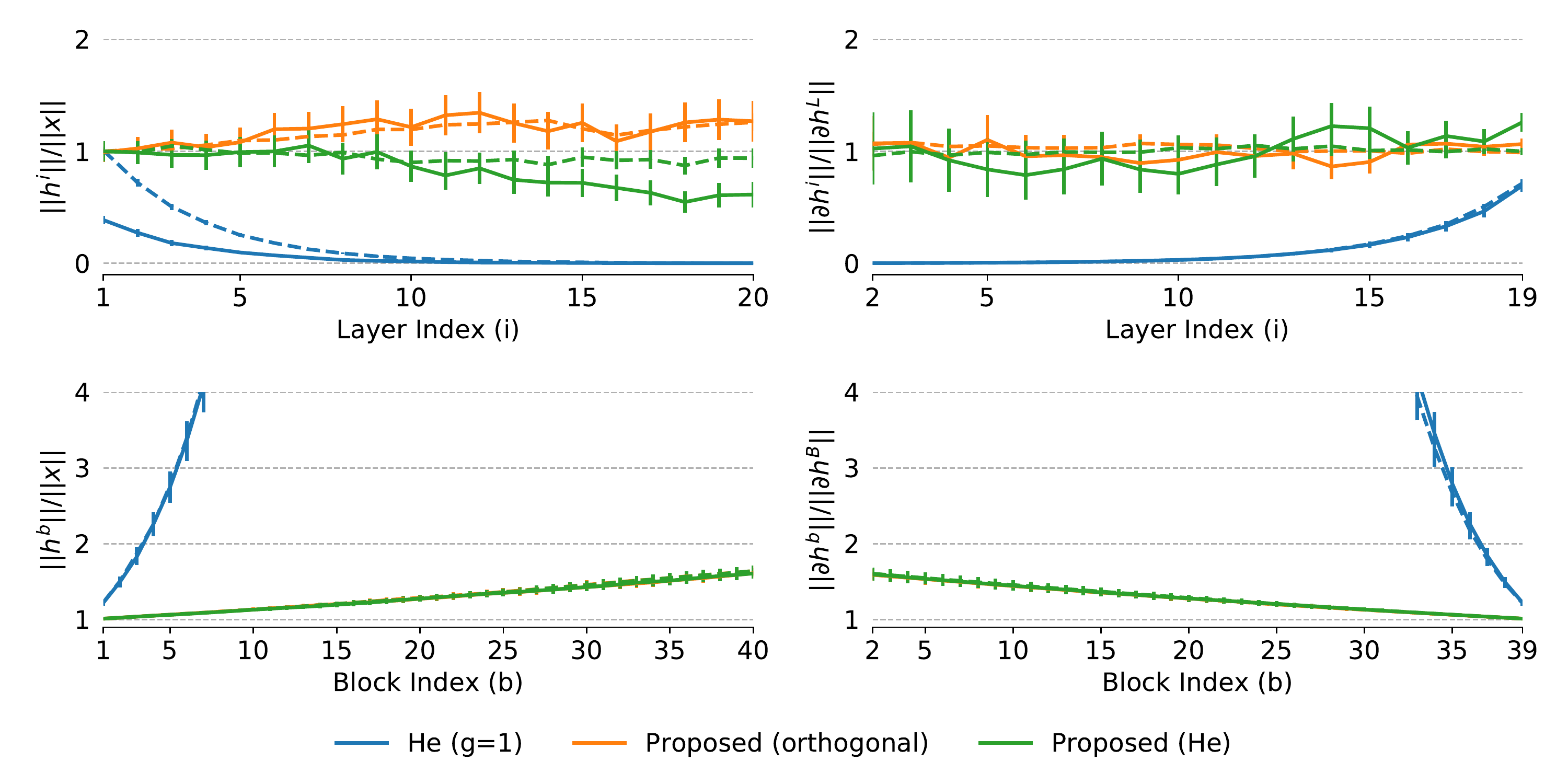}
  \caption{
    Experiments on weight normalized networks using synthetic data to confirm theoretical predictions. \textbf{Top:} feed forward networks. \textbf{Bottom:} residual networks. We report results for networks of $\text{width} \sim \mathcal{U}(150, 250)$ (solid lines) and $\text{width} \sim \mathcal{U}(950, 1050)$ (dashed lines). The proposed initialization prevents explosion/vanishing of the norm of hidden activations (left) and gradients (right) across layers at initialization. For ResNets, norm growth is $\mathcal{O}(1)$ for an arbitrary depth network. Naively initializing $\mathbf{g} = \mathbf{1}$ results in vanishing/exploding signals. 
  }
  \label{fig:WN_toy}
\vspace{-10pt}
\end{figure}

\textbf{Forward pass:}
we first study the forward pass and derive an initialization scheme such that for any given input, the norm of hidden activation of any layer and input norm are asymptotically equal. Failure to do so prevents training to begin, as studied by \citet{hanin2018start} for vanilla deep feedforward networks. 
The theorem below shows that a normalized linear transformation followed by ReLU non-linearity is a norm preserving transform in expectation when proper scaling is used.
\begin{theorem}
\label{theorem_fwd_expected_norm_normed_relu}
Let $\mathbf{v} = ReLU\left(\sqrt{2n/m} \cdot \hat{\mathbf{R}} \mathbf{u}\right)$, where $\mathbf{u} \in \mathbb{R}^n$ and $\hat{\mathbf{R}} \in \mathbb{R}^{m \times n}$. If ${\mathbf{R}}_{i} \stackrel{i.i.d.}{\sim} P$ where $P$ is any isotropic distribution in $\mathbb{R}^n$, or alternatively $\hat{\mathbf{R}}$ is a randomly generated matrix with orthogonal rows, then for any fixed vector $\mathbf{u}$, $\mathbb{E}[\lVert \mathbf{v} \rVert^2] = K_n \cdot \lVert \mathbf{u} \rVert^2$ where,
\begin{align}
    K_n = \begin{cases}
    \frac{2S_{n-1}}{S_n} \cdot \left( \frac{2}{3}\cdot \frac{4}{5}\hdots \frac{n-2}{n-1} \right) & \text{if $n$ is even} \\
    \frac{2S_{n-1}}{S_n} \cdot \left( \frac{1}{2}\cdot \frac{3}{4}\hdots \frac{n-2}{n-1} \right) \cdot \frac{\pi}{2} & \text{otherwise}
    \end{cases}
\end{align}
and $S_n$ is the surface area of a unit $n$-dimensional sphere.
\end{theorem}
The constant $K_n$ seems hard to evaluate analytically, but remarkably, we empirically find that $K_n=1$ for all integers $n>1$. Thus applying the above theorem to Eq.~\ref{eq_norm_relu_layer} implies that every hidden layer in a weight normalized ReLU network is norm preserving for an infinitely wide network if the elements of $\textbf{g}^l$ are initialized with $\sqrt{2n_{l-1}/n_l}$. Therefore, we can recursively apply the above argument to each layer in a normalized deep ReLU network starting from the input to the last layer and have that the network output norm is approximately equal to the input norm, i.e.~$\lVert f_{\mathbf{\theta}}(\mathbf{x}) \rVert \approx \lVert \mathbf{x} \rVert$. Figure~\ref{fig:WN_toy} (top left) shows a synthetic experiment with a 20 layer weight normalized MLP that empirically confirms the above theory. Details for this experiment can be found in the supplementary material.

\textbf{Backward pass:} 
the goal of studying the backward pass is to derive conditions for which gradients do not explode nor vanish, which is essential for gradient descent based training. Therefore, we are interested in the value of $\lVert \frac{\partial \ell(f_{\mathbf{\theta}}(\mathbf{x}), \mathbf{y})}{\partial \mathbf{a}^l} \rVert$ for different layers, which are indexed by $l$. To prevent exploding/vanishing gradients, the value of this term should be similar for all layers. We begin by writing the recursive relation between the value of this derivative for consecutive layers,
\begin{align}
\label{eq_dlda_decomposition_normed_relu}
    \frac{\partial \ell(f_{\mathbf{\theta}}(\mathbf{x}), \mathbf{y})}{\partial \mathbf{a}^l} &= \frac{\partial \mathbf{a}^{l+1}}{\partial \mathbf{a}^{l}} \cdot \frac{\partial \ell(f_{\mathbf{\theta}}(\mathbf{x}), \mathbf{y})}{\partial \mathbf{a}^{l+1}} \\
\label{eq_dlda_decomposition_normed_relu_2}
    &=\textbf{g}^{l+1} \odot \mathbbm{1}(\mathbf{a}^l) \odot \left(  \hat{\mathbf{W}}^{l+1^T}  \frac{\partial \ell(f_{\mathbf{\theta}}(\mathbf{x}), \mathbf{y})}{\partial \mathbf{a}^{l+1}} \right)
\end{align}
We note that conditioned on a fixed $\mathbf{h}^{l-1}$, each dimension of $\mathbbm{1}(\mathbf{a}^l)$ in the above equation follows an i.i.d. sampling from Bernoulli distribution with probability 0.5 at initialization.
This is formalized in Lemma 1 in the supplementary material.  
We now consider the following theorem,
\begin{theorem}
\label{proposition_bwd_norm_preserve_normed_relu_w_normed}
Let $\mathbf{v} = \sqrt{2} \cdot \mathbf{z} \odot \left(\hat{\mathbf{R}}^T \mathbf{u}\right)$, where $\mathbf{u} \in \mathbb{R}^m$, $\mathbf{R} \in \mathbb{R}^{m \times n}$ and $\mathbf{z} \in \mathbb{R}^n$. If each ${\mathbf{R}}_{i} \stackrel{i.i.d.}{\sim} P$ where $P$ is any isotropic distribution in $\mathbb{R}^n$ or alternatively $\hat{\mathbf{R}}$ is a randomly generated matrix with orthogonal rows and $\mathbf{z}_{i} \stackrel{i.i.d.}{\sim} \text{Bernoulli}(0.5)$, then for any fixed vector $\mathbf{u}$, $\mathbb{E}[\lVert \mathbf{v} \rVert^2] = \lVert \mathbf{u} \rVert^2$.
\end{theorem}
In order to apply the above theorem to Eq.~\ref{eq_dlda_decomposition_normed_relu_2}, we assume that $\mathbf{u}:= \frac{\partial \ell(f_{\mathbf{\theta}}(\mathbf{x}), \mathbf{y})}{\partial \mathbf{a}^{l+1}}$ is independent of the other terms, similar to \citet{he2016deep}. This simplifies the analysis by allowing us to treat $ \frac{\partial \ell(f_{\mathbf{\theta}}(\mathbf{x}), \mathbf{y})}{\partial \mathbf{a}^{l+1}}$ as fixed and take expectation w.r.t. the other terms, over $\mathbf{W}^l$ and $\mathbf{W}^{l+1}$. Thus $\lVert \frac{\partial \ell(f_{\mathbf{\theta}}(\mathbf{x}), \mathbf{y})}{\partial \mathbf{a}^l} \rVert = \lVert \frac{\partial \ell(f_{\mathbf{\theta}}(\mathbf{x}), \mathbf{y})}{\partial \mathbf{a}^{l+1}} \rVert$ $\forall l$ if we initialize $\textbf{g}^l =\sqrt{2}\cdot \mathbf{1}$. This also shows that $\frac{\partial \mathbf{a}^{l+1}}{\partial \mathbf{a}^{l}}$ is a norm preserving transform. 
Hence applying this theorem recursively to Eq.~\ref{eq_dlda_decomposition_normed_relu_2} for all $l$ yields that $\lVert \frac{\partial \ell(f_{\mathbf{\theta}}(\mathbf{x}), \mathbf{y})}{\partial \mathbf{a}^l} \rVert \approx \frac{\partial \ell(f_{\mathbf{\theta}}(\mathbf{x}), \mathbf{y})}{\partial \mathbf{a}_L}$ $\forall l$ thereby avoiding gradient explosion/vanishment. 
Note that the above result is strictly better for orthogonal weight matrices compared with other isotropic distributions (see proof). 
Figure~\ref{fig:WN_toy} (top right) shows a synthetic experiment with a 20 layer weight normalized MLP to confirm the above theory. The details for this experiment are provided in the supplementary material.


We also point out that the $\sqrt{2}$ factor that appears in theorems \ref{theorem_fwd_expected_norm_normed_relu} and \ref{proposition_bwd_norm_preserve_normed_relu_w_normed} is due to the presence of ReLU activation. In the absence of ReLU, this factor should be 1. We will use this fact in the next section with the ResNet architecture.

\textbf{Implementation details:}
since there is a discrepancy between the initialization required by the forward and backward pass, we tested both (and combinations of them) in our preliminary experiments and found the one proposed for the forward pass to be superior. We therefore propose to initialize weight matrices $\mathbf{W}^l$ to be orthogonal\footnote{We note that \citet{saxe2013exact} propose to initialize weights of un-normalized deep ReLU networks to be orthogonal with scale $\sqrt{2}$. Our derivation and proposal is for weight normalized ReLU networks where we study both Gaussian and orthogonal initialization and show the latter is superior.}, $\mathbf{b}^l=\mathbf{0}$, and $\mathbf{g}^l=\sqrt{2n_{l-1}/n_l}\cdot \mathbf{1}$, where $n_{l-1}$ and $n_l$ represent the fan-in and fan-out of the $l^{th}$ layer respectively. Our results apply to both fully-connected and convolutional\footnote{For convolutional layers with kernel size $k$ and $c$ channels, we define $n_{l-1}=k^2c_{l-1}$ and $n_{l}=k^2c_{l}$ \citep{he2015delving}.} networks.


\section{Residual Networks}
\label{sec_residual_nets}
Similar to the previous section, we derive an initialization strategy for ResNets in the infinite width setting. We define a residual network $\mathcal{R}(\{ F_b(\cdot) \}_{b=0}^{B-1}, \theta, \alpha)$ with $B$ residual blocks and parameters $\theta$ whose output is denoted as $f_\theta(\cdot) = \mathbf{h}^B$, and the hidden states are defined recursively as,
\begin{align}
\label{eq_resnet}
    \mathbf{h}^{b+1} &:= \mathbf{h}^{b} + \alpha F_b(\mathbf{h}^{b}) \mspace{20mu} b \in \{ 0,1, \ldots, B-1 \}
\end{align}
where $\mathbf{h}^{0} = \mathbf{x}$ is the input, $\mathbf{h}^{b}$ denotes the hidden representation after applying $b$ residual blocks and $\alpha$ is a scalar that scales the output of the $b$-{th} residual blocks. The $b\text{-th} \in \{0,1, \ldots, B-1\}$ residual block $F_b(\cdot)$ is a feed-forward ReLU network. We discuss how to deal with shortcut connections during initialization separately.
We use the notation ${<}\cdot,\cdot{>}$ to denote dot product between the argument vectors.


\textbf{Forward pass}: 
here we derive an initialization strategy for residual networks that prevents information in the forward pass from exploding/vanishing independent of the number of residual blocks, assuming that each residual block is initialized such that it preserves information in the forward pass. 
\begin{theorem}
\label{theorem_residual_net_expected_activation}
Let $\mathcal{R}(\{ F_b(\cdot) \}_{b=0}^{B-1}, \theta, \alpha)$ be a residual network with output $f_\theta(\cdot)$. Assume that each residual block $F_b(.)$ ($\forall b$) is designed such that at initialization, $\lVert F_b(\mathbf{h}) \rVert = \lVert \mathbf{h} \rVert$ for any input $\mathbf{h}$ to the residual block, and ${<}\mathbf{u}, F_b(\mathbf{u}){>} \approx 0$. If we set $\alpha=1/\sqrt{B}$, then $\exists c \in [\sqrt{2},\sqrt{e}]$, such that,
\begin{align}
\label{eq_resnet_fwd}
    \lVert f_\theta(\mathbf{x}) \rVert \approx c \cdot \lVert \mathbf{x} \rVert 
\end{align}
\end{theorem}
The assumption ${<}\mathbf{u}, F_b(\mathbf{u}){>} \approx 0$ is reasonable because at initialization, $F_b(\mathbf{u})$ is a random transformation in a high dimensional space which will likely rotate a vector to be orthogonal to itself. To understand the rationale behind the second assumption, $\lVert F_b(\mathbf{h}) \rVert = \lVert \mathbf{h} \rVert$, recall that $F_b(.)$ is essentially a non-residual network. Therefore we can initialize each such block using the scheme developed in Section~\ref{sec_normed_deep_relu} which due to Theorem \ref{theorem_fwd_expected_norm_normed_relu} (see discussion below it) will guarantee that the norm of the output of $F_b(\cdot)$ equals the norm of the input to the block. Figure \ref{fig:WN_toy} (bottom left) shows a synthetic experiment with a 40 block weight normalized ResNet to confirm the above theory. The ratio of norms of output to input lies in $[\sqrt{2},\sqrt{e}]$ independent of the number of residual blocks exactly as predicted by the theory. The details for this experiment can be found in the supplementary material.

\textbf{Backward pass:}
we now study the backward pass for residual networks. 

\begin{theorem}
\label{theorem_resnet_bwd}
Let $\mathcal{R}(\{ F_b(\cdot) \}_{b=0}^{B-1}, \theta, \alpha)$ be a residual network with output $f_\theta(\cdot)$. Assume that each residual block $F_b(\cdot)$ ($\forall b$) is designed such that at initialization, $\lVert \frac{\partial F_b(\mathbf{h}^b)}{\partial \mathbf{h}^b} \mathbf{u} \rVert = \lVert \mathbf{u} \rVert$ for any fixed input $\mathbf{u}$ of appropriate dimensions, and ${<}\frac{\partial \ell}{\partial \mathbf{h}^b}, \frac{\partial \mathbf{F}_{b-1}}{\partial \mathbf{h}^{b-1}} \cdot \frac{\partial \ell}{\partial \mathbf{h}_{b}}{>} \approx 0$. If $\alpha = \frac{1}{\sqrt{B}}$, then $\exists c \in [\sqrt{2},\sqrt{e}]$, such that,
\begin{align}
    \lVert \frac{\partial \ell}{\partial \mathbf{h}^1} \rVert \approx  c \cdot \lVert \frac{\partial \ell}{\partial \mathbf{h}^{B}} \rVert 
\end{align}
\end{theorem}
The above theorem shows that scaling the output of the residual block with $1/\sqrt{B}$ prevents explosion/vanishing of gradients irrespective of the number of residual blocks. The rationale behind the assumptions is similar to that given for the forward pass above. Figure \ref{fig:WN_toy} (bottom right) shows a synthetic experiment with a 40 block weight normalized ResNet to confirm the above theory. Once again, the ratio of norms of gradient w.r.t. input to output lies in $[\sqrt{2},\sqrt{e}]$ independent of the number of residual blocks exactly as predicted by the theory. The details can be found in the supplementary material.

\textbf{Shortcut connections:}
a ResNet often has $K$ \textit{stages}~\cite{he2016deep}, where each stage is characterized by one \textit{shortcut} connection and $B_k$ residual blocks, leading to a total of $\sum_{k=1}^K{B_k}$ blocks. 
In order to account for shortcut connections, we need to ensure that the input and output of each \textit{stage} in a ResNet are at the same scale; the same argument applies during the backward pass. To achieve this, we scale the output of the residual blocks in each stage using the total number of residual blocks in that stage. Then theorems \ref{theorem_residual_net_expected_activation} and \ref{theorem_resnet_bwd} treat each stage of the network as a ResNet and normalize the flow of information in both directions to be independent of the number of residual blocks.

\textbf{Implementation details:}
we consider ResNets with shortcut connections and architecture design similar to that proposed in \cite{he2016deep} with the exception that our residual block structure is Conv$\rightarrow$ReLU$\rightarrow$Conv, similar to $B(3,3)$ blocks in \citep{zagoruyko2016wide}, as illustrated in the supplementary material\footnote{More generally, our residual block design principle is $D \times$[Conv$\rightarrow$ReLU$\rightarrow$]Conv, where $D \in \mathbb{Z}$.}.
Weights of all layers in the network are initialized to be orthogonal and biases are set to zero. 
The gain parameter of weight normalization is initialized to be 
$\textbf{g} = \sqrt{\gamma \cdot \text{fan-in} / \text{fan-out}} \cdot \mathbf{1}$. 
We set $\gamma = 1/B_k$ for the last convolutional layer of each residual block in the $k$-th stage\footnote{We therefore absorb $\alpha$ in Eq. \ref{eq_resnet} into the gain parameter $\mathbf{g}$.}. For the rest of layers we follow the strategy derived in Section \ref{sec_normed_deep_relu}, with $\gamma=2$ when the layer is followed by ReLU, and $\gamma=1$ otherwise.

\section{Experiments}

We study the impact of initialization on weight normalized networks across a wide variety of configurations. Among others, we compare against the data-dependent initialization proposed by \citet{salimans2016weight}, which initializes $\mathbf{g}$ and $\mathbf{b}$ so that all pre-activations in the network have zero mean and unit variance based on estimates collected from a single minibatch of data. 

Code is publicly available at \url{https://github.com/victorcampos7/weightnorm-init}. 
We refer the reader to the supplementary material for detailed description of the hyperparameter settings for each experiment, as well as for initial reinforcement learning results.

\subsection{Robustness Analysis of Initialization methods-- Depth, Hyper-parameters  and Seed}

The difficulty of training due to exploding and vanishing gradients increases with network depth. In practice, depth often complicates the search for hyperparameters that enable successful optimization, if any. This section presents a thorough evaluation of the impact of initialization on different network architectures for increasing depths, as well as their robustness to hyperparameter configurations. We benchmark fully-connected networks on MNIST~\citep{mnistlecun}, whereas CIFAR-10~\citep{cifar} is considered for convolutional and residual networks. We tune hyperparameters individually for each network depth and initialization strategy on a set of held-out examples, and report results on the test set. We refer the reader to the supplementary material for a detailed description of the considered hyperparameters.

\begin{figure}
\centering
  \includegraphics[width=\textwidth]{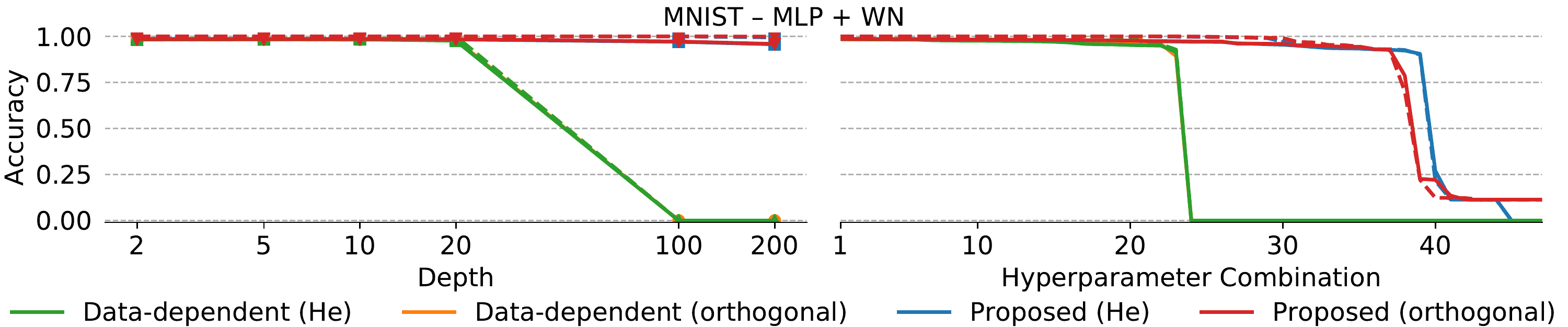}
  \caption{Results for MLPs on MNIST. Dashed lines denote train accuracy, and solid lines denote test accuracy. The accuracy of diverged runs is set to $0$. \textbf{Left:} Accuracy as a function of depth. A held-out validation set is used to select the best model for each configuration. \textbf{Right:} Accuracy for each job in our hyperparameter sweep, depicting robustness to hyperparameter configurations.}
  \label{fig:mnist-mlp-results}
\vspace{-10pt}
\end{figure}

\textbf{Fully-connected networks:} results in Figure~\ref{fig:mnist-mlp-results} (left) show that the data-dependent initialization can be used to train networks of up to depth 20, but training diverges for deeper nets even when using very small learning rates, e.g.~$10^{-5}$. On the other hand, we managed to successfully train very deep networks with up to 200 layers using the proposed initialization. When analyzing all runs in the grid search, we observe that the proposed initialization is more robust to the particular choice of hyperparameters (Figure~\ref{fig:mnist-mlp-results}, right). In particular, the proposed initialization allows using learning rates up to $10\times$ larger for most depths.

\begin{figure}
\centering
  \includegraphics[width=\linewidth]{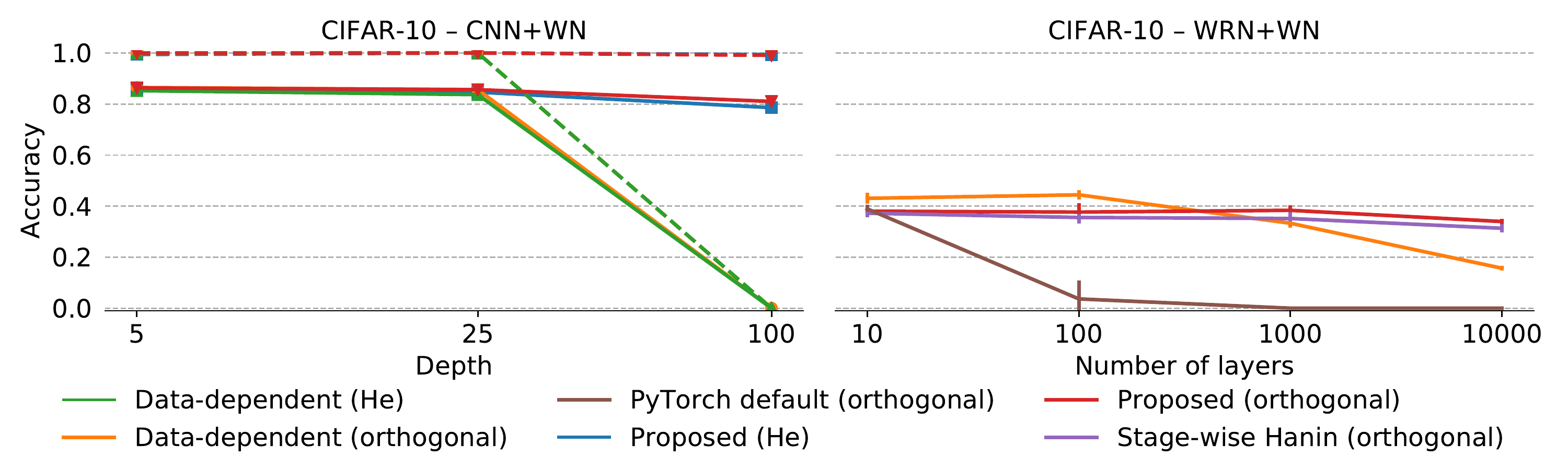}
  \caption{Accuracy as a function of depth on CIFAR-10 for CNNs (\textbf{left}), and WRNs (\textbf{right}). Dashed lines denote train accuracy, and solid lines denote validation accuracy. Note that WRNs are trained for a single epoch due to the computational burden of training extremely deep networks.}
  \label{fig:cifar10-cnn+wrn-acc_v_depth}
\end{figure}

\textbf{Convolutional networks:} we adopt a similar architecture to that in \citet{xiao2018dynamical}, where all layers have $3\times3$ kernels and a fixed width. The two first layers use a stride of $2$ in order to reduce the memory footprint. Results are depicted in Figure~\ref{fig:cifar10-cnn+wrn-acc_v_depth} (left) and show a similar trend to that observed for fully-connected nets, with the data-dependent initialization failing at optimizing very deep networks.

\textbf{Residual networks:} we construct residual networks of varying depths by controlling the number of residual blocks per stage in the wide residual network (WRN) architecture with $k=1$. Training networks with thousands of layers is computationally intensive, so we measure the test accuracy after a single epoch of training~\citep{zhang2019fixup}. 
We consider two additional baselines for these experiments: (1)~the default initialization in PyTorch\footnote{\url{https://pytorch.org/docs/stable/_modules/torch/nn/utils/weight_norm.html}}, which initializes $g_i = \lVert {\mathbf{W}}_i \rVert_2$, and (2)~a modification of the initialization proposed by \citet{hanin2018start} to fairly adapt it to weight normalized multi-stage ResNets. For the $k$-th stage with $B_k$ blocks, the stage-wise Hanin scheme initializes the gain of the last convolutional layer in each block as $\mathbf{g}=0.9^b\mathbf{1}$, where $b\in\{1, \ldots, B_k\}$ refers to the block number within a stage. 
All other parameters are initialized in a way identical to our proposal, so that information across the layers within residual blocks remains preserved.
We report results over 5 random seeds for each configuration in Figure~\ref{fig:cifar10-cnn+wrn-acc_v_depth} (right), which shows that the proposed initialization achieves similar accuracy rates across the wide range of evaluated depths. PyTorch's default initialization diverges for most depths, and the data-dependent baseline converges significantly slower for deeper networks due to the small learning rates used in order to avoid divergence. Despite the stage-wise Hanin strategy and the proposed initialization achieve similar accuracy rates, we were able to use an order of magnitude larger learning rates with the latter, which denotes an increased robustness against hyperparameter configurations. 

\begin{figure}
\centering
  \includegraphics[width=\linewidth]{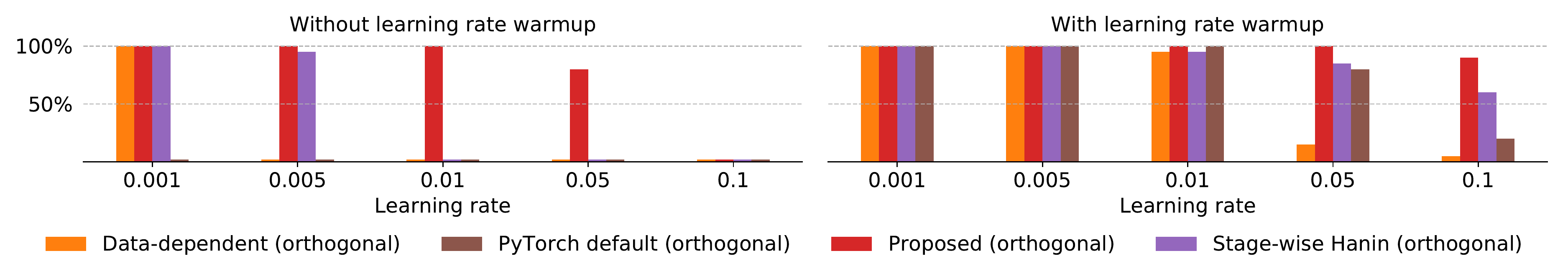}
  \caption{Robustness to seed of different initialization schemes on WRN-40-10. We launch 20 training runs for every configuration, and measure the percentage of runs that reach epoch 3 without diverging. Weight normalized ResNets benefit from learning rate warmup, which enables the usage of higher learning rates. The proposed initialization is the most robust scheme across all configurations.}
  \label{fig:cifar10-wrn-robustness}
\vspace{-15pt}
\end{figure}

To further evaluate the robustness of each initialization strategy, we train WRN-40-10 networks for 3 epochs with different learning rates, with and without learning rate warmup \citep{goyal2017accurate}. We repeat each experiment 20 times using different random seeds, and report the percentage of runs that successfully completed all 3 epochs without diverging in Figure~\ref{fig:cifar10-wrn-robustness}. We observed that learning rate warmup greatly improved the range of learning rates that work well for all initializations, but the proposed strategy manages to train more robustly across all tested configurations.

\subsection{Comparison with Batch Normalization}

Existing literature has pointed towards an implicit regularization effect of batch normalization~\citep{luo2018towards}, which prevented weight normalized models from matching the final performance of batch normalized ones~\citep{gitman2017comparison}. 
On the other hand, previous works have shown that larger learning rates facilitate finding wider minima which correlate with better generalization performance \citep{keskar2016large,jastrzkebski2017three,smith2018bayesian,smith2017don}, and the proposed initialization and learning rate warmup have proven very effective in stabilizing training for high learning rates.
This section aims at evaluating the final performance of weight normalized networks trained with high learning rates, and compare them with batch normalized networks.


We evaluate models on CIFAR-10 and CIFAR-100. We set aside 10\% of the training data for hyperparameter tuning, whereas some previous works use the test set for such purpose~\citep{he2016deep,zagoruyko2016wide}. This difference in the experimental setup explains why the achieved error rates are slightly larger than those reported in the literature. For each architecture we use the default hyperparameters reported in literature for batch normalized networks, and tune only the initial learning rate for weight normalized models.

Results in Table~\ref{table:sota_results} show that the proposed initialization scheme, when combined with learning rate warmup, allows weight normalized residual networks to achieve comparable error rates to their batch normalized counterparts.
We note that previous works reported a large generalization gap between weight and batch normalized networks~\citep{shang2017exploring,gitman2017comparison}.
The only architecture for which the batch normalized variant achieves a superior performance is WRN-40-10, for which the weight normalized version is not able to completely fit the training set before reaching the epoch limit. This phenomena is different to the generalization gap reported in previous works, and might be caused by sub-optimal learning rate schedules that were tailored for networks with batch normalization.

\begin{table}[ht]
    \vspace{-12pt}
    \centering
    \caption{Comparison between Weight Normalization with proposed initialization and Batch Normalization. Results are reported as mean and std over 5 runs.}
    \label{table:sota_results}
    \vspace{0.15cm}
    \begin{tabular}{llll}
    \toprule
    \textbf{Dataset} &\textbf{Architecture} & \textbf{Method} & {\textbf{Test Error ($\%$)}}\\
    \midrule
    \multirow{13}{*}{CIFAR-10} & \multirow{4}{*}{ResNet-56}& 
        WN w/ datadep init & 9.19 $\pm$ 0.24\\
        && WN w/ proposed init & 7.87 $\pm$ 0.14\\
        && WN w/ proposed init + warmup & 7.20 $\pm$ 0.12\\
        && BN (\citet{he2016deep})& 6.97\\
    \cmidrule{2-4}
    &\multirow{5}{*}{ResNet-110} & 
        WN w/ datadep init & 9.33 $\pm$ 0.10\\
        && WN w/ proposed init & 7.71 $\pm$ 0.14\\
        && WN w/ proposed init + warmup & 6.69 $\pm$ 0.11\\
        && WN (\citet{shang2017exploring})  & 7.46\\
        && BN (\citet{he2016deep})& 6.61 $\pm$ 0.16\\
    \cmidrule{2-4}
    & \multirow{4}{*}{WRN-40-10} & 
        WN w/ datadep init + cutout & 6.10 $\pm$ 0.23\\
        && WN w/ proposed init + cutout & 4.74 $\pm$ 0.14\\
        && WN w/ proposed init + cutout + warmup  & 4.75 $\pm$ 0.08\\
        && BN w/ orthogonal init + cutout & 3.53 $\pm$ 0.38\\
    \midrule
    \multirow{4}{*}{CIFAR-100} & \multirow{4}{*}{ResNet-164} & 
        WN w/ datadep init + cutout & 30.26 $\pm$ 0.51 \\
        && WN w/ proposed init + cutout & 27.30 $\pm$ 0.49\\
        && WN w/ proposed init + cutout + warmup & 25.31 $\pm$ 0.26 \\
        && BN w/ orthogonal init + cutout & 25.52 $\pm$ 0.17\\
    \bottomrule
    \end{tabular}
\vspace{-10pt}
\end{table} 

\subsection{Initialization Method and Generalization Gap}
The motivation behind designing good parameter initialization is mainly for better optimization at the beginning of training, and it is not apparent why our initialization is able to reduce the generalization gap between weight normalized and batch normalized networks~\cite{gitman2017comparison,shang2017exploring}. On this note we point out that a number of papers have shown how using stochastic gradient descent (SGD) with larger learning rates facilitate finding wider minima which correlate with better generalization performance~\cite{keskar2016large,jastrzkebski2017three,smith2018bayesian,smith2017don}. Additionally, it is often not possible to use large learning rates with weight normalization with traditional initializations. Therefore we believe that the use of large learning rate allowed by our initialization played an important role in this aspect. In order to understand why our initialization allows using large learning rates compared with existing ones, we compute the (log) spectral norm of the Hessian at initialization (using Power method) for the various initialization methods considered in our experiments using $10\%$ of the training samples. They are shown in Table~\ref{table:hessian}. We find that the local curvature (spectral norm) is smallest for the proposed initialization. These results are complementary to the seed robustness experiment shown in Figure~\ref{fig:cifar10-wrn-robustness}. 

\begin{table}[ht]
\vspace{-12pt}
\begin{center}
\caption{Log (base 10) spectral norm of Hessian at
  initialization for different initializations. Smaller values imply lower curvature. \textit{N/A} means that the computation diverged. The proposed strategy initializes at a point with lowest curvature, which explains why larger learning rates can be used.}
\label{table:hessian}
\vspace{0.15cm}
\resizebox{\linewidth}{!}{
    \begin{tabular}{lccccc}
        \toprule
        \textbf{Dataset}& \textbf{Model}  & \textbf{PyTorch default}  & \textbf{Data-dependent} & \textbf{Stage-wise Hanin}  & \textbf{Proposed} \\
        \midrule
        CIFAR-10 & WRN-40-10 & 4.68 $\pm$ 0.60 & 3.01 $\pm$ 0.02 & 7.14 $\pm$ 0.72 & 1.31 $\pm$ 0.12\\
        \midrule
        CIFAR-100 & ResNet-164 & 9.56 $\pm$ 0.54 & 2.68 $\pm$ 0.09 & N/A & 1.56 $\pm$ 0.18\\
        \bottomrule
    \end{tabular}
  }
  \end{center}
\vspace{-15pt}
\end{table}

\section{Conclusion and Future Work}
Weight normalization (WN) is frequently used in different network architectures due to its simplicity. However, the lack of existing theory on parameter initialization of weight normalized networks has led practitioners to arbitrarily pick existing initializations designed for un-normalized networks. To address this issue, we derived parameter initialization schemes for weight normalized networks, with and without residual connections, that avoid explosion/vanishment of information in the forward and backward pass. To the best of our knowledge, no prior work has formally studied this setting. 
Through thorough empirical evaluation, we showed that the proposed initialization increases robustness to network depth, choice of hyper-parameters and seed compared to existing initialization methods that are not designed specifically for weight normalized networks. 
We found that the proposed scheme initializes networks in low curvature regions, which enable the use of large learning rates. 
By doing so, we were able to significantly reduce the performance gap between batch and weight normalized networks which had been previously reported in the literature. Therefore, we hope that our proposal replaces the current practice of choosing arbitrary initialization schemes for weight normalized networks.

We believe our proposal can also help in achieving better performance using WN in settings which are not well-suited for batch normalization. One such scenario is training of recurrent networks in backpropagation through time settings, which often suffer from exploding/vanishing gradients, and batch statistics are timestep-dependent~\cite{cooijmans2016recurrent}. The current analysis was done for feedforward networks, and we plan to extend it to the recurrent setting. Another application where batch normalization often fails is reinforcement learning, as good estimates of activation statistics are not available due to the online nature of some of these algorithms. We confirmed the benefits of our proposal in preliminary reinforcement learning experiments, which can be found in the supplementary material.

\section*{Acknowledgments}
We would like to thank Giancarlo Kerg for proofreading the paper. DA was supported by IVADO during his time at Mila where part of this work was done.

\bibliography{ref}
\bibliographystyle{plainnat}

\newpage
\clearpage
\appendix 


\setcounter{page}{1}
\appendix
\onecolumn

\section{Experimental setup}
\subsection{Details about Figure 1 (top)}
We use a weight normalized 20 layer MLP with 1000 randomly generated input samples in $\mathbf{R}^{500}$. We test three initialization strategies. (1)~He initialization \citep{he2016deep} for the weight matrices and the gain parameter $\mathbf{g}$ for all layers are initialized to 1. (2)~Proposed initialization, where weights are initialized to be orthogonal and gains are set as $\sqrt{2n_{l-1}/n_l}$. (3)~Proposed initialization, where weights are initialized using He initialization and gains are set as $\sqrt{2n_{l-1}/n_l}$. In all cases biases are set to 0. At initialization itself, we forward propagate the 1000 randomly generated input samples, measure the norm of hidden activations, and compute the mean and standard deviation of the ratio of norm of hidden activation to the norm of the input. This is shown in Figure 1 (top left). In Figure 1 (top right), we similarly record the norm of hidden activation gradient by backpropagating 1000 random error vectors, and measure the ratio of the norm of hidden activation gradient to the norm of the error vector. We find that the proposed initialization preserves norm in both directions while vanilla He initialization fails. This shows the importance of proper initialization of the $\gamma$ parameter of weight normalization.

\subsection{Details about Figure 1 (bottom)}
We use a weight normalized ResNet with 40 residual blocks with 1000 randomly generated input samples in $\mathbf{R}^{500}$. The network architecture is exactly as described in Eq. 6, with a residual block composed of two fully connected (FC) layers, i.e.~FC1 $\rightarrow$ ReLU $\rightarrow$ FC2. The weight normalization layers are inserted after FC layers. We test three initialization strategies. (1)~He initialization~\cite{he2016deep} for all the weight matrices, and gain parameter $\mathbf{g} = \mathbf{1}$. (2)~Proposed initialization where weights are initialized to be orthogonal and gains are set as $\sqrt{2 \cdot \text{fan-in}/\text{fan-out}}$ for FC1 and $\sqrt{\text{fan-in}/(40 \cdot \text{fan-out})}$ for FC2. (3)~Proposed initialization where weights are initialized using He initialization and gains are set same as in the previous case. In all cases biases are set to 0. At initialization itself, we forward propagate the 1000 randomly generated input samples, measure the norm of hidden activations $\mathbf{h}^b$ and compute the mean and standard deviation of the ratio of norm of hidden activation to the norm of the input $\mathbf{x}$. This is shown in Figure 1 (bottom left). In Figure 1 (bottom right), we similarly record the norm of hidden activation gradient by backpropagating 1000 random error vectors and measure the ratio of the norm of hidden activation gradient $\frac{\partial \ell}{\partial \mathbf{h}^b}$ to the norm of the error vector $\frac{\partial \ell}{\partial \mathbf{h}^{B}}$. We find that the proposed initialization preserves norm in both directions while vanilla He initialization fails. This shows the importance of proper initialization of the $\mathbf{g}$ parameter of weight normalization.


\begin{table}[ht]
\begin{center}
  \caption{Hyperparameters for MNIST experiments. Values between brackets were used in the grid search. Learning rate of $0.00001$ was considered for depths $100$ and $200$ only.}
  \label{table:mnist_mlp_hyperparams}
  \vspace{0.15cm}
    \begin{tabular}{ll}
      \toprule
      \textbf{Parameter}	        & \textbf{Value} \\
      \midrule
      Data split                    & 10\% of the original train is set aside for validation purposes\\
      Number of hidden layers		& $\{ 2, 5, 10, 20, 100, 200 \}$ \\
      Size of hidden layers	        & $\{ 512, 1024 \}$ \\
      Number of epochs	            & $ 150 $ \\
      Initial learning rate	        & $\{ 0.1, 0.01, 0.001, 0.0001, 0.00001^* \}$ \\
      Learning rate schedule	    & Decreased by $10\times$ at epochs $50$ and $100$ \\
      Batch size                    & $128$ \\
      Weight decay	                & $0.0001$ \\
      Optimizer 	                & SGD with $\text{momentum}=0.9$ \\
      \bottomrule
    \end{tabular}
  \end{center}
\end{table}

\begin{table}[ht]
\begin{center}
  \caption{Hyperparameters for CNN experiments on CIFAR-10. Values between brackets were used in the grid search. Learning rate of $0.001$ was considered for depth $100$ only.}
  \label{table:cifar10_cnn_hyperparams}
  \vspace{0.15cm}
    \begin{tabular}{ll}
      \toprule
      \textbf{Parameter}	        & \textbf{Value} \\
      \midrule
      Data split                    & 10\% of the original train is set aside for validation purposes\\
      \multirow{4}{*}{Architecture} & $2 \times [\text{Conv2D } 3\times3/2, 512]$  \\
                                    & $(\text{N}-2) \times [\text{Conv2D } 3\times3/1, 512]$ \\
                                    & Global Average Pooling \\
                                    & 10-d Linear, softmax \\
      Number of hidden layers (N)	& $\{ 5, 25, 100 \}$ \\
      Number of epochs	            & $ 500 $ \\
      Initial learning rate	        & $\{ 0.01, 0.001^* \}$ \\
      Learning rate schedule	    & Decreased by $10\times$ at epoch $166$ \\
      Batch size                    & $100$ \\
      Weight decay	                & $\{ 0.001, 0.0001 \}$ \\
      Optimizer 	                & SGD without momentum \\
      \bottomrule
    \end{tabular}
  \end{center}
\end{table}

\newcommand{\resblock}[2]{
  #2$\times$\(\left[
      \begin{array}{c}
        \text{Conv2D 3$\times$3, #1}\\[-.1em]
        \text{ReLU}\\[-.1em]
        \text{Conv2D 3$\times$3, #1}
      \end{array}
    \right]\)
}
\newcommand{\convsize}[1]{#1$\times$#1}
\newcommand{\convname}[1]{#1}
\def\cellheight{0.50cm}
\begin{figure}[ht] 
  \begin{tabular}{cc}
        \centering
        \begin{tabular}{ccc}
            \toprule
            Group name & Output size & Block type \\
            \midrule
            conv1 & \convsize{32} & [Conv2D 3$\times$3, 16$\times$k] \\
            \convname{conv2} & \convsize{32} & \resblock{16$\times$k}{N} \\[\cellheight]
            \convname{conv3} & \convsize{16} & \resblock{32$\times$k}{N} \\[\cellheight]
            \convname{conv4} & \convsize{8} & \resblock{64$\times$k}{N} \\
            out & \convsize{1} & [average pooling, 10-d fc, softmax] \\
            \bottomrule
        \end{tabular}
        &
        \begin{subfigure}{0.25\textwidth}
        \centering
        \smallskip
          \includegraphics[width=\linewidth]{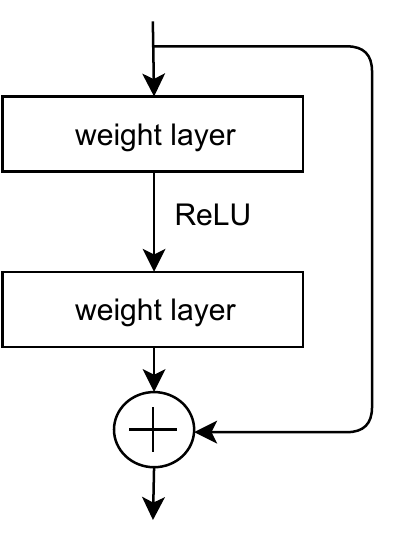}
          \label{fig:resblock}
        \end{subfigure}
        \\
  \end{tabular}
  \vspace{0.1cm}
  \caption{\textbf{Left:} Architecture of Wide Residual Networks considered in this work. Downsampling is performed through strided convolutions by the first layers in groups \texttt{conv3} and \texttt{conv4}. \textbf{Right:} Structure of a residual block. Note that there is no non-linearity after residual conections, unlike \cite{he2016deep}.}
  \label{table:wrn_architecture}
\end{figure}

\begin{table}[ht]
\begin{center}
  \caption{Hyperparameters for WRN experiments on CIFAR-10. Values between brackets were used in the grid search. Learning rates smaller than $10^{-5}$ were considered for $N=1666$ (10,000 layers) only.}
  \label{table:cifar10_wrn_hyperparams}
  \vspace{0.15cm}
 \resizebox{\linewidth}{!}{
    \begin{tabular}{ll}
      \toprule
      \textbf{Parameter}	                & \textbf{Value} \\
      \midrule
      Data split                            & 10\% of the original train is set aside for validation purposes\\
      WRN's $N$ (residual blocks per stage)   & $\{ 1, 16, 166, 1666 \}$ \\
      WRN's $k$ (width factor)                & $ 1 $ \\
      Number of epochs	                    & $ 1 $ \\
      Initial learning rate	                & $\{7\cdot10^{-1}, 3\cdot10^{-1}, 1\cdot10^{-1}, \ldots, 10^{-5}, \ldots, 10^{-7*}\}$ \\
      Batch size                            & $128$ \\
      Weight decay	                        & $0.0005$ \\
      Optimizer 	                        & SGD with $\text{momentum}=0.9$ \\
      \bottomrule
    \end{tabular}
 }
  \end{center}
\end{table}

\section{Reinforcement Learning experiments}
\label{sec:rl_experiments}

Despite its tremendous success in supervised learning applications, Batch Normalization~\cite{ioffe2015batch} is seldom used in reinforcement learning (RL), as the online nature of some of the methods and the strong correlation between consecutive batches hinder its performance. These properties suggest the need for normalization techniques like Weight Normalization~\cite{salimans2016weight}, which are able to accelerate and stabilize training of neural networks without relying on minibatch statistics.

We consider the Asynchronous Advantage Actor Critic (A3C) algorithm~\cite{mnih2016asynchronous}, which maintains a policy and a value function estimate which are updated asynchronously by different workers collecting experience in parallel. Updates are estimated based on $n$-step returns from each worker, resulting in highly correlated batches of $n$ samples, whose impact is mitigated through the asynchronous nature of updates. This setup is not well suited for Batch Normalization and, to the best of our knowledge, no prior work has successfully applied it to this type of algorithm.

\begin{figure}[h]
\centering
  \includegraphics[width=\linewidth]{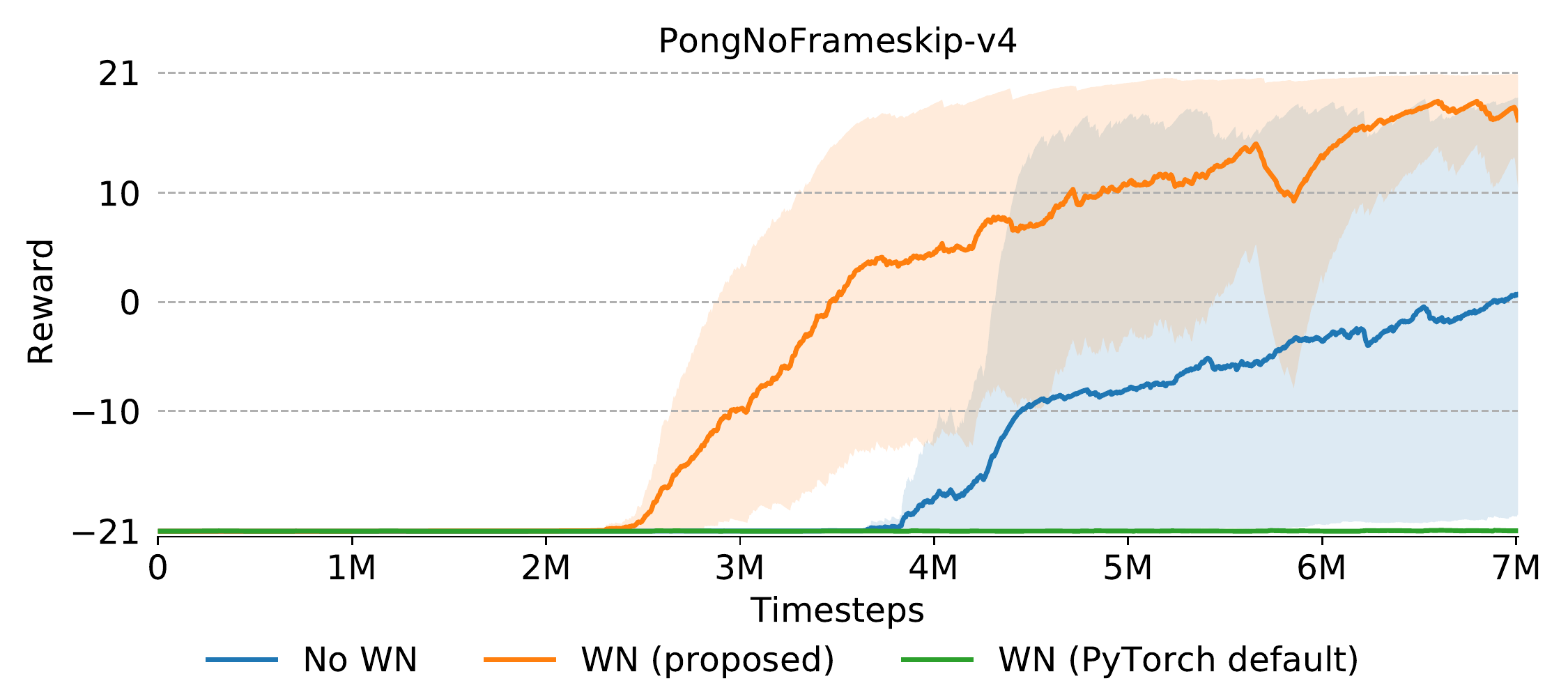}
  \caption{Learning progress in Pong. Shading shows maximum and minimum over 3 random seeds, while dark lines indicate the mean. Weight normalization with the proposed initialization improves convergence speed and reduces variance across seeds. These results highlight the importance of initialization in weight normalized networks, as using the default initialization in PyTorch prevents training to start.}
  \label{fig:rl_pong}
\end{figure}

We evaluate agents using Atari environments in the Arcade Learning Environment~\cite{bellemare2013arcade}. Our initial experiments with the deep residual architecture introduced by \citet{espeholt2018impala} show that adding Weight Normalization improves convergence speed and robustness to hyperparameter configurations across different environments. However, we did not observe important differences between initialization schemes for these weight normalized models. Despite being significantly deeper than previous architectures used in RL, this model is still relatively shallow for supervised learning standards, and we observed in our computer vision experiments that performance differences arise for deeper architectures or high learning rates. The latter is known to cause catastrophic performance degradation in deep RL due to excessively large policy updates~\cite{schulman2015trust}, so we opt for building a much deeper residual network with 100 layers. Collecting experience with such a deep policy is a very slow process even when using GPU workers. Given this computational burden, we use hyperparameters tuned in initial experiments for the deep network introduced by \citet{espeholt2018impala}, and report initial results in one of the simplest environments\footnote{Collecting 7M timesteps of experience took approximately 10h on a single GPU shared by 6 workers. Even though this amount of experience is enough to solve Pong, A3C usually needs many more interactions to learn competitive policies in more complex environments.} in Figure~\ref{fig:rl_pong}.

We observe that the weight normalized policy with the proposed initialization manages to solve the task much faster than the un-normalized architecture. Perhaps surprisingly, the weight normalized policy with the sub-optimal initialization is not able to solve the environment in the given timestep budget, and it performs even worse than the un-normalized policy. These results highlight the importance of proper initialization even when using normalization techniques.

The deep network architecture considered in this experiment is excessively complex for the considered task, which can be solved with much smaller networks. However, with the development of ever complex environments~\cite{vinyals2017starcraft} and distributed learning algorithms that can take advantage of massive computational resources~\cite{espeholt2018impala}, recent results have shown that RL can benefit from techniques that have found success in the supervised learning community, such as deeper residual networks~\cite{silver2017mastering,espeholt2018impala}. The aforementioned findings suggest that RL applications could benefit from techniques that help training very deep networks robustly in the future.

\section{Proofs}
\label{app_proofs}

\begin{theorem}
Let $\mathbf{v} = ReLU\left(\sqrt{\frac{2n}{m}} \cdot \hat{\mathbf{R}} \mathbf{u}\right)$, where $\mathbf{u} \in \mathbb{R}^n$ and $\hat{\mathbf{R}} \in \mathbb{R}^{m \times n}$. If ${\mathbf{R}}_{i} \stackrel{i.i.d.}{\sim} P$ where $P$ is any isotropic distribution in $\mathbb{R}^n$, or alternatively $\hat{\mathbf{R}}$ is a randomly generated matrix with orthogonal rows, then for any fixed vector $\mathbf{u}$, $\mathbb{E}[\lVert \mathbf{v} \rVert^2] = K_n \cdot \lVert \mathbf{u} \rVert^2$ where,
\begin{align}
    K_n = \begin{cases}
    \frac{2S_{n-1}}{S_n} \cdot \left( \frac{2}{3}\cdot \frac{4}{5}\hdots \frac{n-2}{n-1} \right) & \text{if $n$ is odd} \\
    \frac{2S_{n-1}}{S_n} \cdot \left(\frac{1}{2}\cdot  \frac{3}{4}\hdots \frac{n-2}{n-1} \right) \cdot \frac{\pi}{2} & \text{otherwise}
    \end{cases}
\end{align}
and $S_n$ is the surface are of a unit $n$-dimensional sphere.

\textbf{Proof}: During the proof, take note of the distinction between the notations $\hat{\mathbf{R}}_{i}$ and ${\mathbf{R}}_{i}$. Our goal is to compute,
\begin{align}
    \mathbb{E}[\lVert \mathbf{v} \rVert^2] &= \mathbb{E}[\sum_{i=1}^m v_i^2]\\
    &= \sum_{i=1}^m \mathbb{E}[ v_i^2]
\end{align}
Suppose the weights are randomly generated to be orthogonal with uniform probability over all rotations. Due to the linearity of expectation, when taking the expectation of any unit $v_i$ over the randomly generated orthogonal weight matrix, the expectation marginalizes over all the rows of the weight matrix except the $i^{th}$ row. As a consequence, for each unit $i$, the expectation is over an isotropic random variable since the orthogonal matrix is generated randomly with uniform probability over all rotations. Therefore, we can equivalently write,
\begin{align}
    \mathbb{E}[\lVert \mathbf{v} \rVert^2] &=  m \mathbb{E}[ v_i^2]
\end{align}
Note that the above equality would trivially hold if all rows of the weight matrix were sampled i.i.d. from an isotropic distribution. In other words, the above equality holds irrespective of the two choice of distributions used for sampling the weight matrix.

We have,
\begin{align}
    \mathbb{E} [v_i^2] &= \mathbb{E} [\max(0, \sqrt{\frac{2n}{m}} \cdot \hat{\mathbf{R}}_{i}^T\mathbf{u})^2]\\
    &= \int_{\mathbf{R}_{i}} p({\mathbf{R}}_{i})\max(0, \sqrt{\frac{2n}{m}} \cdot \lVert \mathbf{u} \rVert \cos \theta)^2
\end{align}
where $p({\mathbf{R}}_{i})$ denotes the probability distribution of the random variable ${\mathbf{R}}_{i}$, and $\theta$ is the angle between vectors $\hat{\mathbf{R}}_{i}$ and $\mathbf{u}$. Hence $\theta$ is a function of $\hat{\mathbf{R}}_{i}$. Since $\mathbf{R}_{i}$ is sampled from an isotropic distribution, the direction and scale of $\mathbf{R}_{i}$ are independent. Thus,
\begin{align}
    \int p({\mathbf{R}}_{i})\max(0, \sqrt{\frac{2n}{m}} \cdot \lVert \mathbf{u} \rVert \cos \theta)^2 &= \int_{\mathbf{R}_{i}} p({\lVert \mathbf{R}}_{i} \rVert) \int_{\hat{\mathbf{R}}_{i}} p(\hat{\mathbf{R}}_{i})\max(0, \sqrt{\frac{2n}{m}} \cdot\lVert \mathbf{u} \rVert \cos \theta)^2\\
    &= \int_{\hat{\mathbf{R}}_{i}} p(\hat{\mathbf{R}}_{i})\max(0, \sqrt{\frac{2n}{m}} \cdot\lVert \mathbf{u} \rVert \cos \theta)^2\\
    &= {\frac{2n}{m}} \cdot \lVert \mathbf{u} \rVert ^2 \int_{\hat{\mathbf{R}}_{i}} p(\hat{\mathbf{R}}_{i})\max(0,\cos \theta)^2
\end{align}
Since $P$ is an isotropic distribution in $\mathbb{R}^n$, the likelihood of all directions is uniform. It essentially means that $p(\hat{\mathbf{R}}_{i})$ can be seen as a uniform distribution over the surface area of a unit $n$-dimensional sphere. We can therefore re-parameterize $p(\hat{\mathbf{R}}_{i})$ in terms of $\theta$ by aggregating the density $p(\hat{\mathbf{R}}_{i})$ over all points on this $n$-dimensional sphere at a fixed angle $\theta$ from the vector $\mathbf{u}$. This is similar to the idea of Lebesgue integral. To achieve this, we note that all the points on the $n$-dimensional sphere at a constant angle $\theta$ from $\mathbf{u}$ lie on an $(n-1)$-dimensional sphere of radius $\sin \theta$. Thus the aggregate density at an angle $\theta$ from $\mathbf{u}$ is the ratio of the surface area of the $(n-1)$-dimensional sphere of radius $\sin \theta$ and the surface area of the unit $(n)$-dimensional sphere. Therefore,
\begin{align}
    \int_{\hat{\mathbf{R}}_{i}} p(\hat{\mathbf{R}}_{i})\max(0,\cos \theta)^2 &= \int_0^\pi \frac{S_{n-1}}{S_n} \cdot |\sin^{n-1} \theta| \cdot \max(0,\cos \theta)^2\\
    &= \frac{S_{n-1}}{S_n} \int_0^{\pi/2}  \sin^{n-1} \theta \cos^2 \theta\\
    &= \frac{S_{n-1}}{S_n} \int_0^{\pi/2}  \sin^{n-1} \theta (1 - \sin^2 \theta)\\
    &= \frac{S_{n-1}}{S_n} \int_0^{\pi/2}  \sin^{n-1} \theta  - \sin^{n+1} \theta
\end{align}
Now we use a known result in existing literature that uses integration by parts to evaluate the integral of exponentiated sine function, which states,
\begin{align}
    \int \sin^n \theta = -\frac{1}{n} \sin^{n-1}\theta \cos \theta + \frac{n-1}{n}\int \sin^{n-2} \theta
\end{align}
Since our integration is between the limits $0$ and $\pi/2$, we find that the first term on the R.H.S. in the above expression is 0. Recursively expanding the $n-2^{th}$ power sine term, we can similarly eliminate all such terms until we are left with the integral of $\sin \theta$ or $\sin^0 \theta$ depending on whether $n$ is odd or even. For the case when $n$ is odd, we get,
\begin{align}
    \int_0^{\pi/2} \sin^n \theta &= \left( \frac{2}{3}\cdot \frac{4}{5}\hdots \frac{n-1}{n} \right)\int_0^{\pi/2} \sin \theta\\
    &= -\left( \frac{2}{3}\cdot \frac{4}{5}\hdots \frac{n-1}{n} \right) \cos \theta |_0^{\pi/2}\\
    &= \left( \frac{2}{3}\cdot \frac{4}{5}\hdots \frac{n-1}{n} \right)
\end{align}

For the case when $n$ is even, we similarly get,
\begin{align}
    \int_0^{\pi/2} \sin^n \theta &= \left( \frac{1}{2}\cdot \frac{3}{4}\cdot \frac{5}{6}\hdots \frac{n-1}{n} \right)\int_0^{\pi/2} \sin^0 \theta\\
    &= \left( \frac{1}{2}\cdot \frac{3}{4}\cdot \frac{5}{6}\hdots \frac{n-1}{n} \right)\int_0^{\pi/2} 1 \\
    &= \left( \frac{1}{2}\cdot \frac{3}{4}\cdot \frac{5}{6}\hdots \frac{n-1}{n} \right) \cdot \frac{\pi}{2}
\end{align}

Thus,
\begin{align}
    \int_0^{\pi/2}  \sin^{n-1} \theta  - \sin^{n+1} \theta = \begin{cases}
    \frac{1}{n} \cdot \left( \frac{2}{3}\cdot \frac{4}{5}\hdots \frac{n-2}{n-1} \right) & \text{if $n$ is odd} \\
    \frac{1}{n} \cdot \left( \frac{1}{2}\cdot \frac{3}{4}\hdots \frac{n-2}{n-1} \right) \cdot \frac{\pi}{2} & \text{otherwise}
    \end{cases}
\end{align}
Define,
\begin{align}
    K_n = \begin{cases}
    \frac{2S_{n-1}}{S_n} \cdot \left( \frac{2}{3}\cdot \frac{4}{5}\hdots \frac{n-2}{n-1} \right) & \text{if $n$ is odd} \\
    \frac{2S_{n-1}}{S_n} \cdot \left( \frac{1}{2}\cdot \frac{3}{4}\hdots \frac{n-2}{n-1} \right) \cdot \frac{\pi}{2} & \text{otherwise}
    \end{cases}
\end{align}
Then,
\begin{align}
    \int_{\hat{\mathbf{R}}_{i}} p(\hat{\mathbf{R}}_{i})\max(0,\cos \theta)^2 &=  \frac{0.5K_n}{n}
\end{align}
Thus,
\begin{align}
    \mathbb{E}[\lVert \mathbf{v} \rVert^2] &= m \mathbb{E} [v_i^2]\\
    &= m \cdot \frac{2n}{m} \cdot \lVert \mathbf{u} \rVert^2 \cdot \frac{0.5K_n}{n}\\
    &= K_n \cdot \lVert \mathbf{u} \rVert^2
\end{align}

which proves the claim. $\square$
\end{theorem}

\begin{lemma}
\label{proposition_preactivation_properties}
If network weights are sampled i.i.d. from a Gaussian distribution with mean 0 and biases are 0 at initialization, then conditioned on $\mathbf{h}^{l-1}$, each dimension of $ \mathbbm{1}(\mathbf{a}^l)$ follows an i.i.d. Bernoulli distribution with probability 0.5 at initialization.

\textbf{Proof}: Note that $\mathbf{a}^l := \mathbf{W}^l \mathbf{h}^{l-1}$ at initialization (biases are 0) and $\mathbf{W}^l$ are sampled i.i.d. from a random distribution with mean 0. Therefore, each dimension $\mathbf{a}^l_i$ is simply a weighted sum of i.i.d. zero mean Gaussian, which is also a 0 mean Gaussian random variable.

To prove the claim, note that the indicator operator applied on a random variable with 0 mean and symmetric distribution will have equal probability mass on both sides of 0, which is the same as a Bernoulli distributed random variable with probability 0.5. Finally, each dimension of $\mathbf{a}^l$ is i.i.d. simply because all the elements of $\mathbf{W}^l$ are sampled i.i.d., and hence each dimension of $\mathbf{a}^l$  is a weighted sum of a different set of i.i.d. random variables. $\square$
\end{lemma}

\begin{theorem}
Let $\mathbf{v} = \sqrt{2} \cdot \left(\hat{\mathbf{R}}^T \mathbf{u}\right) \odot \mathbf{z}$, where $\mathbf{u} \in \mathbb{R}^m$, $\mathbf{R} \in \mathbb{R}^{m \times n}$ and $\mathbf{z} \in \mathbb{R}^n$. If each ${\mathbf{R}}_{i} \stackrel{i.i.d.}{\sim} P$ where $P$ is any isotropic distribution in $\mathbb{R}^n$ or alternatively $\hat{\mathbf{R}}$ is a randomly generated matrix with orthogonal rows and $\mathbf{z}_{i} \stackrel{i.i.d.}{\sim} \text{Bernoulli}(0.5)$, then for any fixed vector $\mathbf{u}$, $\mathbb{E}[\lVert \mathbf{v} \rVert^2] = \lVert \mathbf{u} \rVert^2$.

\textbf{Proof}: Our goal is to compute,
\begin{align}
    \mathbb{E}[\lVert \mathbf{v} \rVert^2] &= {2} \cdot \mathbb{E}[ \lVert (\sum_{i=1}^n \hat{\mathbf{R}}_i u_i) \odot \mathbf{z} \lVert^2  ]\\
    &= {2} \cdot \mathbb{E}[  \sum_{j=1}^m (\sum_{i=1}^n \hat{\mathbf{R}}_{ij} u_i)^2 \cdot {z}_j^2   ]\\
    &= {2} \cdot \mathbb{E}[{z}_j^2] \cdot \mathbb{E}[ \sum_{j=1}^m (\sum_{i=1}^n \hat{\mathbf{R}}_{ij} u_i)^2  ]\\
    &=  \mathbb{E}[ \lVert (\sum_{i=1}^n \hat{\mathbf{R}}_i u_i) \lVert^2  ]\\
    &=  \mathbb{E}[  \sum_{i=1}^n  u_i^2 \lVert \hat{\mathbf{R}}_i \rVert^2 + \sum_{i \neq j} u_i u_j \cdot \hat{\mathbf{R}}_i^T\hat{\mathbf{R}}_j   ]\\
    &=    \lVert \mathbf{u} \rVert^2 + \sum_{i \neq j} u_i u_j \cdot \mathbb{E}[ \hat{\mathbf{R}}_i^T\hat{\mathbf{R}}_j   ]\\
    &=    \lVert \mathbf{u} \rVert^2 + \sum_{i \neq j} u_i u_j \cdot \mathbb{E}[ \cos \phi   ]
\end{align}
where $\phi$ is the angle between $\hat{\mathbf{R}}_i$ and $\hat{\mathbf{R}}_j$. For orthogonal matrix $\hat{\mathbf{R}}$ $\cos \phi$ is always 0, while for $\hat{\mathbf{R}}$ such that each ${\mathbf{R}}_{i} \stackrel{i.i.d.}{\sim} P$ where $P$ is any isotropic distribution, $\mathbb{E}[ \cos \phi   ]=0$. Thus for both cases\footnote{This also suggests that orthogonal initialization is strictly better than Gaussian initialization since the result holds without the dependence on expectation in contrast to the Gaussian case.} we have that,
\begin{align}
    \mathbb{E}[\lVert \mathbf{v} \rVert^2] = \lVert \mathbf{u} \rVert^2
\end{align}
which proves the claim. $\square$
\end{theorem}

\begin{theorem}
Let $\mathcal{R}(\{ F_b(.) \}_{b=0}^{B-1}, \theta, \alpha)$ be a residual network with output $f_\theta(.)$. Assume that each residual block $F_b(.)$ ($\forall b$) is designed such that at initialization, $\lVert F_b(\mathbf{h}) \rVert = \lVert \mathbf{h} \rVert$ for any input $\mathbf{h}$ to the residual block, and $<\mathbf{h}, F_b(\mathbf{h})> \approx 0$. If we set $\alpha=1/\sqrt{B}$, then,
\begin{align}
    \lVert f_\theta(\mathbf{x}) \rVert^2 \approx c \cdot \lVert \mathbf{x} \rVert^2 
\end{align}
where $c \in [\sqrt{2},\sqrt{e}]$.

\textbf{Proof}: Let $\mathbf{x}$ denote the input of the residual network. Consider the first hidden state $\mathbf{h}^1$ given by,
\begin{align}
    \mathbf{h}^1 &:= \mathbf{x} + \alpha F_1(\mathbf{x})
\end{align}
Then the squared norm of $\mathbf{h}^1$ is given by,
\begin{align}
    \lVert \mathbf{h}^1 \rVert^2 &= \lVert \mathbf{x} + \alpha F_1(\mathbf{x}) \rVert^2 \\
    &= \lVert \mathbf{x} \rVert^2 + \alpha^2 \lVert F_1(\mathbf{x}) \rVert^2 + 2\alpha <\mathbf{x}, F_1(\mathbf{x})>
\end{align}
Since $\lVert \mathbf{F}_1(\mathbf{x}) \rVert^2 = \lVert \mathbf{x} \rVert^2$ and $<\mathbf{x}, F_1(\mathbf{x})> \approx 0$ due to our assumptions, we have,
\begin{align}
    \lVert \mathbf{h}^1 \rVert^2 \approx \lVert \mathbf{x} \rVert^2 \cdot (1 + \alpha^2) 
\end{align}
Similarly,
\begin{align}
    \mathbf{h}^2 &:= \mathbf{h}^1 + \alpha F_2(\mathbf{h}^1)
\end{align}
Thus,
\begin{align}
    \lVert \mathbf{h}^2 \rVert^2 &= \lVert \mathbf{h}^1 \rVert^2 + \alpha^2 \lVert F_2(\mathbf{h}^1) \rVert^2 + 2\alpha <\mathbf{h}^1, F_2(\mathbf{h}^1)>
\end{align}
Then due to our assumptions we get,
\begin{align}
    \lVert \mathbf{h}^2 \rVert^2 \approx \lVert \mathbf{h}^1 \rVert^2 \cdot (1 + \alpha^2)
\end{align}
Thus we get,
\begin{align}
     \lVert \mathbf{h}^2 \rVert^2 \approx \lVert \mathbf{x} \rVert^2 \cdot (1 + \alpha^2)^2
\end{align}
Extending such inequalities to the $B^{th}$ residual block, we get,
\begin{align}
     \lVert \mathbf{h}^B \rVert^2 \approx \lVert \mathbf{x} \rVert^2 \cdot (1 + \alpha^{2})^B
\end{align}
Setting $\alpha=1/\sqrt{B}$, we get,
\begin{align}
     \lVert \mathbf{h}^B \rVert^2 \approx \lVert \mathbf{x} \rVert^2 \cdot \left(1 + \frac{1}{{B}} \right)^{B}
\end{align}
Note that the factor $\left(1 + \frac{1}{{B}} \right)^{B} \rightarrow e$ as $B \rightarrow \infty$ due to the following well known result,
\begin{align}
    \lim_{B \rightarrow \infty} \left(1 + \frac{1}{B} \right)^B = {e}
\end{align}
Since $B \in \mathbb{Z}$, $\left(1 + \frac{1}{B} \right)^{B/2} $ lies in $[\sqrt{2},\sqrt{e}]$. 

Thus we have proved the claim. $\square$
\end{theorem}

\begin{theorem}
Let $\mathcal{R}(\{ F_b(.) \}_{b=0}^{B-1}, \theta, \alpha)$ be a residual network with output $f_\theta(.)$. Assume that each residual block $F_b(.)$ ($\forall b$) is designed such that at initialization, $\lVert \frac{\partial F_b(\mathbf{h}^b)}{\partial \mathbf{h}^b} \mathbf{u} \rVert = \lVert \mathbf{u} \rVert$ for any fixed input $\mathbf{u}$ of appropriate dimensions, and $<\frac{\partial \ell}{\partial \mathbf{h}^b}, \frac{\partial \mathbf{F}_{b-1}}{\partial \mathbf{h}^{b-1}} \cdot \frac{\partial \ell}{\partial \mathbf{h}_{b}}> \approx 0$. If $\alpha = \frac{1}{\sqrt{B}}$, then,
\begin{align}
    \lVert \frac{\partial \ell}{\partial \mathbf{h}^1} \rVert \approx  c \cdot \lVert \frac{\partial \ell}{\partial \mathbf{h}^{B}} \rVert 
\end{align}
where $c \in [\sqrt{2},\sqrt{e}]$.

\textbf{Proof}: Recall,
\begin{align}
    \mathbf{h}^b &:= \mathbf{x} + \alpha F_b(\mathbf{h}^{b-1})
\end{align}
Therefore, taking derivative on both sides,
\begin{align}
    \frac{\partial \ell}{\partial \mathbf{h}^{b-1}} &= (\mathbf{I} + \alpha \cdot \frac{\partial F_{b}}{\partial \mathbf{h}^{b-1}}) \cdot \frac{\partial \ell}{\partial \mathbf{h}^{b}}\\ 
    &=  \frac{\partial \ell}{\partial \mathbf{h}^{b}} + \alpha \cdot \frac{\partial F_{b}}{\partial \mathbf{h}^{b-1}} \cdot \frac{\partial \ell}{\partial \mathbf{h}^{b}}
\end{align}
Taking norm on both sides,
\begin{align}
    \lVert \frac{\partial \ell}{\partial \mathbf{h}^{b-1}} \rVert^2 = \lVert \frac{\partial \ell}{\partial \mathbf{h}^{b}} \rVert^2 + \alpha^2 \cdot \lVert \frac{\partial F_{b}}{\partial \mathbf{h}^{b-1}} \cdot \frac{\partial \ell}{\partial \mathbf{h}^{b-1}} \rVert^2 + 2\alpha \cdot <\frac{\partial \ell}{\partial \mathbf{h}^{b}},\frac{\partial F_{b}}{\partial \mathbf{h}^{b-1}} \frac{\partial \ell}{\partial \mathbf{h}^{b-1}}  >
\end{align}
Due to our assumptions, we have,
\begin{align}
    \lVert \frac{\partial \ell}{\partial \mathbf{h}^{b-1}} \rVert^2 &\approx \lVert \frac{\partial \ell}{\partial \mathbf{h}^{b}} \rVert^2 + \alpha^2 \cdot \lVert \frac{\partial \ell}{\partial \mathbf{h}^{b-1}} \rVert^2\\
    &= (1+\alpha^2) \cdot \lVert \frac{\partial \ell}{\partial \mathbf{h}^{b-1}} \rVert^2
\end{align}
Applying this result to all $B$ residual blocks we have that,
\begin{align}
    \lVert \frac{\partial \ell}{\partial \mathbf{h}^{1}} \rVert^2 &\approx  (1+\alpha^2)^B \cdot \lVert \frac{\partial \ell}{\partial \mathbf{h}^{B}} \rVert^2
\end{align}
Setting $\alpha=1/\sqrt{B}$, we get,
\begin{align}
    \lVert \frac{\partial \ell}{\partial \mathbf{h}^{1}} \rVert^2 &\approx  (1+1/B)^B \cdot \lVert \frac{\partial \ell}{\partial \mathbf{h}^{B}} \rVert^2
\end{align}
Note that the factor $\left(1 + \frac{1}{{B}} \right)^{B} \rightarrow e$ as $B \rightarrow \infty$ due to the following well known result,
\begin{align}
    \lim_{B \rightarrow \infty} \left(1 + \frac{1}{B} \right)^B = {e}
\end{align}
Since $B \in \mathbb{Z}$, $\left(1 + \frac{1}{B} \right)^{B/2} $ lies in $[\sqrt{2},\sqrt{e}]$. 

Thus we have proved the claim. $\square$
\end{theorem}

\end{document}